\documentclass[letterpaper]{article} 
\usepackage{aaai24}  
\usepackage{times}  
\usepackage{helvet}  
\usepackage{courier}  
\usepackage[hyphens]{url}  
\usepackage{graphicx} 
\usepackage{natbib}  
\usepackage{caption} 
\usepackage{algorithm}
\usepackage{algorithmic}

\usepackage{multirow}
\usepackage{makecell}
\usepackage{amsmath}
\usepackage{array}
\usepackage{booktabs}

\def\0{{\bf 0}}
\def\1{{\bf 1}}

\def\eg{{\em e.g.}}
\def\ie{{\em i.e.}}

\usepackage{newfloat}
\usepackage{listings}
\DeclareCaptionStyle{ruled}{labelfont=normalfont,labelsep=colon,strut=off} 
\floatstyle{ruled}
\newfloat{listing}{tb}{lst}{}
\floatname{listing}{Listing}
\title{Revisiting Graph-Based Fraud Detection in Sight of Heterophily and Spectrum}
\author{
    Fan Xu\textsuperscript{\rm 3},
    Nan Wang\textsuperscript{\rm 1}\footnotemark[1],
    Hao Wu\textsuperscript{\rm 3},
    Xuezhi Wen\textsuperscript{\rm 1},
    Xibin Zhao\textsuperscript{\rm 2}\thanks{Corresponding author.},
    Hai Wan\textsuperscript{\rm 2}
}

\affiliations{
    \textsuperscript{\rm 1} School of Software, Beijing Jiaotong University, Beijing, China \\
    \textsuperscript{\rm 2} BNRist, KLISS, School of Software, Tsinghua University, Beijing, China \\
    \textsuperscript{\rm 3} IAT, University of Science and Technology of China, Hefei, China \\
    \{markxu, wuhao2022\}@mail.ustc.edu.cn;\\
    \{wangnanbjtu, 22126399\}@bjtu.edu.cn; \{zxb, wanhai\}@tsinghua.edu.cn
}

\usepackage{bibentry}

\begin{document}

\maketitle

\begin{abstract}
Graph-based fraud detection (GFD) can be regarded as a challenging semi-supervised node binary classification task. In recent years, Graph Neural Networks (GNN) have been widely applied to GFD, characterizing the anomalous possibility of a node by aggregating neighbor information. However, fraud graphs are inherently heterophilic, thus most of GNNs perform poorly due to their assumption of homophily. In addition, due to the existence of heterophily and class imbalance problem, the existing models do not fully utilize the precious node label information. To address the above issues, this paper proposes a semi-supervised GNN-based fraud detector SEC-GFD. This detector includes a hybrid filtering module and a local environmental constraint module, the two modules are utilized to solve heterophily and label utilization problem respectively. The first module starts from the perspective of the spectral domain, and solves the heterophily problem to a certain extent. Specifically, it divides the spectrum into various mixed-frequency bands based on the correlation between spectrum energy distribution and heterophily. Then in order to make full use of the node label information, a local environmental constraint module is adaptively designed. The comprehensive experimental results on four real-world fraud detection datasets denote that SEC-GFD outperforms other competitive graph-based fraud detectors. We release our code at https://github.com/Sunxkissed/SEC-GFD.
\end{abstract}

\section{Introduction}
With the rapid development of the Internet, various fraudulent entities emerge in fraud activities, and the huge losses caused by this have attracted continuous attention from industry, academia and state agencies~\cite{ref1,ref2,ref3,ref4}. For instance, fraudulent behaviors aimed at online payment have caused huge property losses to clients; and fraudulent or malicious comments on websites may bring negative effects on merchants and consumers.

In recent years, graph-based fraud detection methods~\cite{ref5,ref6} have been widely used in practical applications. Using entities in fraud scenarios as nodes and interactions between entities as edges, graph-level fraud detection can be achieved. Furthermore, based on the constructed fraud graph, as a deep neural network for graph representation learning, Graph Neural Network (GNN)~\cite{ref41} has been demonstrated to be effective in this task. Specifically, GNN employs neural modules to aggregate neighbor information and update node representations, enabling the discrimination of fraudulent entities.

\begin{figure}[!t]
\centering
\includegraphics[width=3.0in]{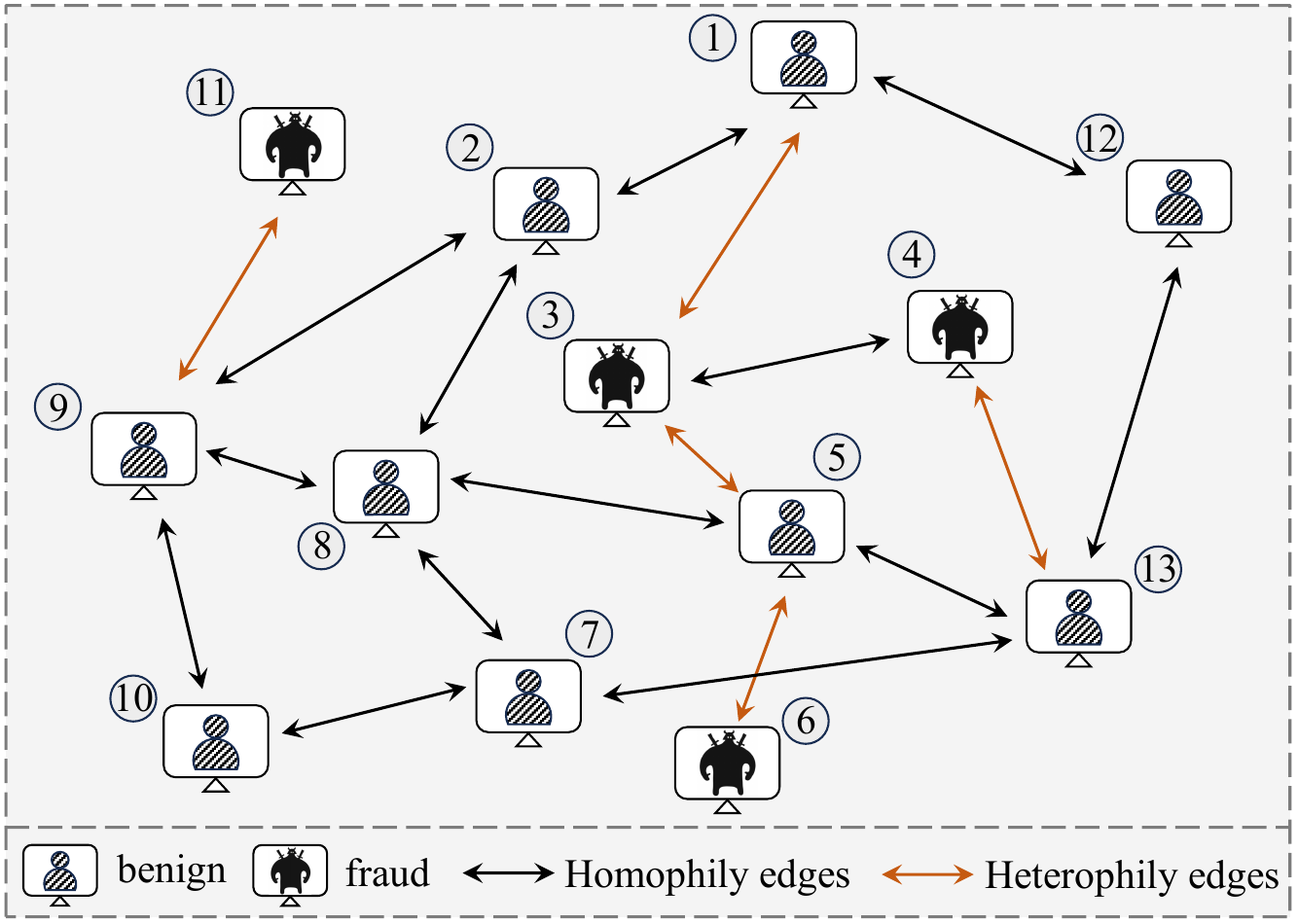}
\caption{Schematic diagram of fraud relationship in real scenarios.}
\label{sample}
\end{figure}

However, fraud detection scenarios are inherently heterophily~\cite{ref7}, \ie, target nodes (like node 5 in Figure~\ref{sample}) and their direct neighbors are prone to different class labels or features. In practical applications, fraudulent entities may use benign entities as springboards to perform high-order interactions with multi-hop neighbors to accomplish fraudulent actions or transmit fraudulent information, \eg, fraud node 3 and 6 are two-hop neighbors in Figure~\ref{sample}. For instance, in financial transaction networks~\cite{ref39}, fraudsters often use non-fraudulent users to conduct transactions.

Due to the existence of heterophily, the performance of conventional graph neural network is unsatisfactory. This is because vanilla GNN is naturally a low-pass filter, which forces the representation of adjacent nodes to be similar during the information aggregation process. Then it may destroy the discriminative information of anomaly nodes. Furthermore, in the domain of fraud detection, heterophily graph datasets exhibit highly imbalanced class distribution properties. This phenomenon results in anomaly nodes being submerged in normal nodes, which manifests as high homophily of normal nodes and high heterophily of anomaly nodes. This is also called the common camouflage problem~\cite{ref9} in fraud detection scenarios. Moreover, current semi-supervised fraud detection methods based on Graph Neural Networks (GNNs) suffer from limitations in terms of label utilization. These approaches typically learn node representations through a message passing mechanism, extract node embeddings from the final layer of the network, and incorporate label information during the training process. Evidently, these approaches overlook the intricate high-order associations between node label information, node characteristics, and their contextual surroundings.

The current paramount concern entails addressing the heterophily issue within the context of extreme category imbalance. The advanced methods can be divided into two types. The first type of the methods is to design a new message aggregation mechanism~\cite{ref38} to simultaneously or separately consider the characteristics of homophily and heterophily edges. The second type of the methods is to identify the heterophily edges on fraud graphs and then perform pruning. According to recent researches, scholars found the spectral `right-shift' phenomenon~\cite{ref24} on both synthetic and real datasets, proving that the higher the anomaly ratio, the more high-frequency components on the graph spectrum. In addition, researchers also demonstrate that, when using the ground truth labels as signals, the proportion of heterophily edges is positively correlated with the spectral energy distribution~\cite{ref25}. Based on these, we conclude that heterophily is not universally harmful, instead it ought to respond distinctively when meeting different filters. In order to verify our conjecture, we conduct sufficient experiments, and the specific related discussions are placed in experiments.

In order to design a GNN-based fraud detector that adapts to high heterophily and has remarkable label utilization, we first propose a conjecture that heterophily edge clipping responds differently to different filters, and then conduct comprehensive experiments to verify the conjecture. Based on the results, we design a \textbf{S}pectrum-\textbf{E}nhanced and \textbf{E}nvironment-\textbf{C}onstrainted \textbf{G}raph \textbf{F}raud \textbf{D}etector (SEC-GFD) to comprehensively integrate rich spectrum and label information into a fraud detector. The framework of the proposed method is shown in Figure~\ref{framework}, consisting of two different modules that address the aforementioned two challenges respectively. The hybrid-pass filter module employs a sophisticated message passing technique utilizing the decomposed spectrum to acquire comprehensive high-frequency representations. The local environmental constraint module enhances the higher-order connections between target nodes and their surrounding environments, thereby mitigating the issue of limited label utilization. The main contributions of this method are summarized as follows:

\begin{itemize}
    \item To address the challenge of heterophily, this paper employs spectrum analysis to partition the graph spectrum into hybrid frequency bands. Then a novel hybrid spectral filter is designed to conduct message passing on these frequency bands separately and finally aggregate.
    \item With the objective of enhancing label utilization, we introduce a local environmental constraint module to assist the training procedure.
    \item We conduct extensive experiments on four real-world benchmark datasets, and the results demonstrate the superior effectiveness and robustness of our proposed SEC-GFD over state-of-the-art methods.
\end{itemize}

\section{Related Work}

\subsubsection{Semi-supervised Node Classification.}
Semi-supervised node classification is one of the most important tasks in graph learning. In recent years, GNNs have achieved excellent performance in semi-supervised node classification, like GCN~\cite{ref10}, GraphSAGE~\cite{ref11}, GAT~\cite{ref12}, and GIN~\cite{ref13}. Compared with traditional graph embedding methods such as DeepWalk~\cite{ref14} and Node2vec~\cite{ref15}, which mainly focus on graph structure, GNN can consider both node features and graph structure information. Each layer of GNN first performs message passing according to the node neighbors, and then updates the node representation through the aggregated information.

\begin{figure*}[!t]
\centering
\includegraphics[width=6.8in]{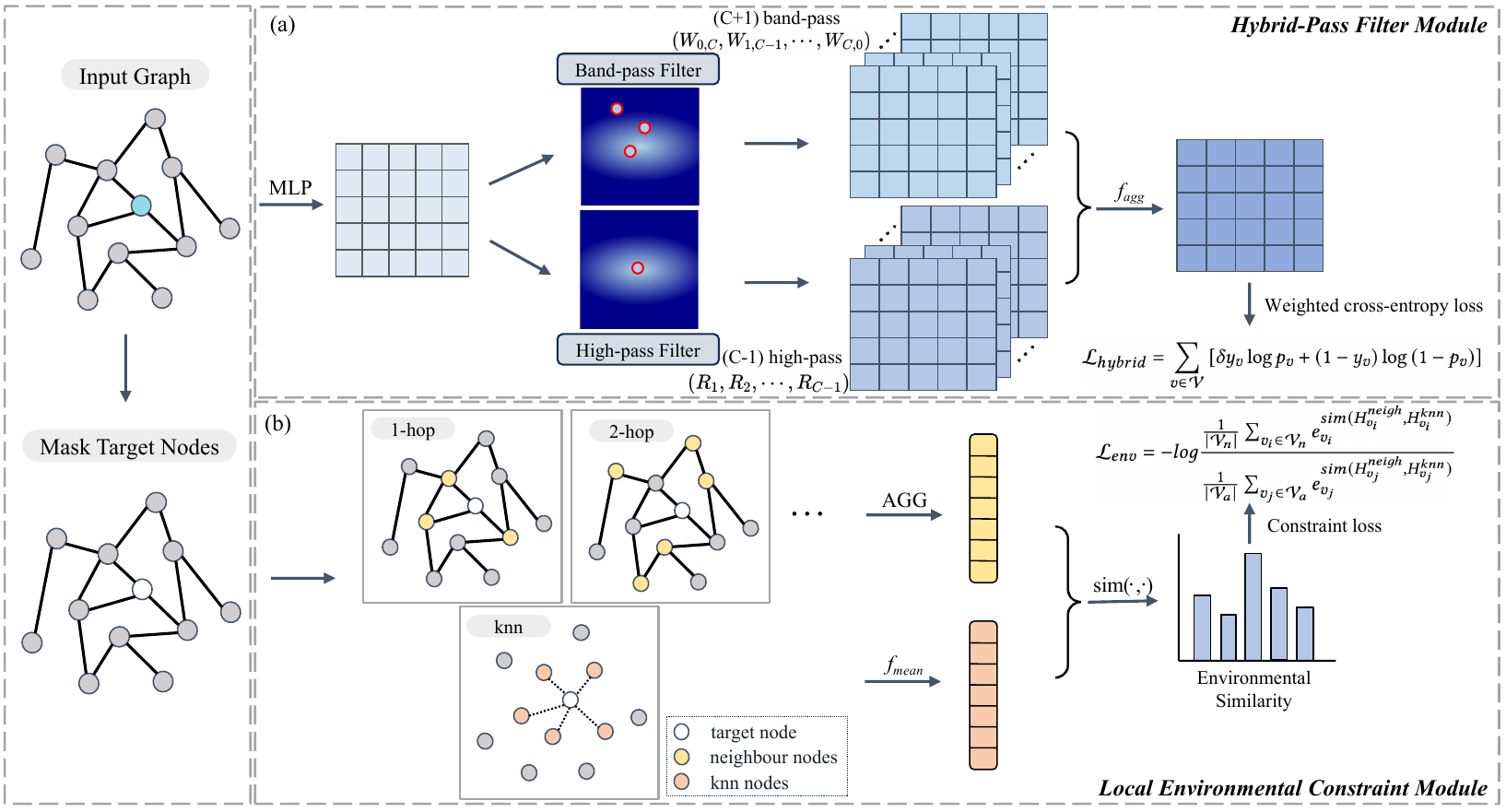}
\caption{The above image presents an overview of our model SEC-GFD, where (a) and (b) respectively demonstrate the details of the hybrid-pass filter module and local environmental constraint module.}
\label{framework}
\end{figure*}

\subsubsection{Popular GNN-based Fraud Detection Model.} 

GNN-based fraud detection (GFD) can be defined as an imbalanced binary classification task that identifies outlier nodes deviating significantly from the majority of nodes. This paper targets node-level fraud detection on static graphs. For instance, GraphConsis~\cite{ref16} combines contextual embeddings with node features and introduces node-related relationship attention weights. CARE-GNN~\cite{ref17} aims at the camouflage problem in fraud detection, and designs GNN enhanced neighbor selection module. PC-GNN~\cite{ref18} designs adaptive subgraph and neighbor sampling methods to fraud detection scenarios. GAGA~\cite{ref19} uses group aggregation to generate discriminative neighborhood information to handle low homophily problems and enhances the original feature space with learnable encoding. This brings some inspiration to our work. GTAN~\cite{ref20} designs gated temporal attention networks to learn node transaction representations.

\vspace{-0.1cm}
\subsubsection{GNN Specially Designed for Heterophily.}

In fraud detection domain, nodes with different labels are often connected, which greatly limits the performance of vanilla GNNs, resulting in the emergence of a large number of studies based on heterophily. 

For instance, GPRGNN~\cite{ref21} combines adaptive Generalized PageRank (GPR) schemes with GNNs to adapt to heterophily graphs. ACM~\cite{ref22} studies heterophily from the perspective of node similarity after aggregation, and then defines a new homophily index. H2-FDetector~\cite{ref23} identifies homophily and heterophily connections through known labeled nodes, and designs different aggregation strategies for them. BWGNN~\cite{ref24} designs a band-pass filter kernel function to transmit the information of each frequency band separately, thus ensuring that the spectrum energy distribution is consistent before and after message passing. GHRN~\cite{ref25} judges and cuts heterophily edges from the perspective of graph spectrum.

\section{Methodology}

This section begins by providing a problem definition and introducing the concepts of graph spectrum and heterophily. Subsequently, it presents the framework of the proposed model, elaborating on each module individually.


\subsection{Problem Definition}
We first introduce the notations of this paper, and then give the formal problem definition. Let $G = (\mathcal{V}, \mathcal{E}, X, Y)$ denote an attributed fraud graph, where $\mathcal{V}=\{v_1,\cdots,v_N\}$ denotes its node set, $\mathcal{E}$ is the edge set, and $A\in\{0,1\}^{N\times N}$ is the adjacency matrix of $G$ with $A_{ij}=1$ if there is a connection between node $v_{i}$ and $v_{j}$. $X\in R^{N\times d}$ denotes the matrix of node attributes where $x_i\in R^d$ is the attribute vector of $v_{i}$, and $Y$ is the corresponding labels of the nodes.

We always formulate the graph-based fraud detection (GFD) as a semi-supervised node-level binary classification problem. Most of the time, anomalies are regarded as positive with label 1, while normal nodes are viewed as negative with label 0~\cite{ref26}. The nodes of the whole fraud graph are divided into the labeled ones and the unlabeled ones, the former are labeled with $Y_{train}$ and the labels of latter ones $Y_{test}$ are invisible during training period. Thus GFD aims to learn an anomaly labeling function for our unlabeled nodes with all known information:

\vspace{-0.3cm}
\begin{equation}
    \setlength{\abovedisplayskip}{3pt}
    \setlength{\belowdisplayskip}{3pt}
    Y_{test} = f(X, A, Y_{train}).
\end{equation}

\subsection{Graph Spectrum and Heterophily}
\subsubsection{Graph Spectrum.}
As $A$ is the adjacency matrix of $G$, we let the graph laplacian matrix $L$ be defined as $D-A$ or as $I-D^{-1/2}AD^{-1/2}$~\cite{ref10}, where $I$ is an identify matrix. $L$ is a positive semidefinite matrix and can be rewritten as $U \Lambda U^{T}$, where $\Lambda$ is a diagonal matrix consisted of eigenvalues $\lambda _{1}\le \lambda _{2}\le\ ... \le \lambda _{N}$ and $U$ is composed of the corresponding $N$ eigenvectors $U=(u_{1},u_{2},...,u_{N})$~\cite{ref29}. 

According to theory of graph signal processing, we can treat the eigenvectors of normalized Laplacian matrix $U$ as the bases in Graph Fourier Transform. Suppose $X=[x_{1},x_{2},...,x_{N}]$ are signals on the graph, we treat $U^{T}X$ as the graph fourier transform~\cite{ref27} of signal $X$. Thus, the convolution between the signal $x$ and convolution kernel $g$ is as follows:

\begin{equation}
    g * x = U((U^T g) \odot (U^T x)) = U g_w U^T x,
\end{equation}

\noindent where $\odot$ is the element-wise product of vectors and $g_w$ is a diagonal matrix representing the convolutional kernel in the spectral domain. Most popular subsequent spectral GNN methods mainly focus on improving the design of $g_w$. And for signals $X$, the Rayleigh quotient represents signals' standardized variance fraction~\cite{ref28}:

\begin{equation}
    R(L, X) = \frac{X^T L X}{X^T X} \\
    =\frac{{\sum_{(i, j)\in \mathcal E}^{}} (x_{i}-x_{j})^2}{{2 \sum_{i \in \nu }^{}} x_{i}^2}.
\end{equation}

Moreover, as discussed in BWGNN~\cite{ref24}, the Rayleigh quotient can also represent the proportion of high-frequency component energy in the spectrum, which matches the overhead variance fraction. The proof details can be found in Appendix A.

\vspace{-0.2cm}
\subsubsection{Heterophily.}
Given a set of labeled nodes and the connections that exist between them, an edge is called a heterophilic connection if its source node and destination node have different labels (i.e., anomaly and normal). For each node, the sum of its heterophily and homophily degree is equal to 1. The heterophily of the node and the whole graph can be defined separately as:

\vspace{-0.3cm}
\begin{equation}
\begin{split}
    hetero(v)&=\frac{1}{|\mathcal{N}(v)|}|\{u:u\in\mathcal{N}(v),y_{u}\neq y_{v}\}|, \\
    hetero({G})&=\frac{1}{|{\mathcal E}|} |\{\varepsilon:  y_{src} \ne y_{dst} \}|,
\end{split}
\end{equation}

\noindent where $|\mathcal N(v)|$ is the number of neighbours of node $v$, and $|{\mathcal E}|$ is the total number of edges in the whole graph, $y_{src}$ and $y_{dst}$ are the labels of the source and destination node of the edge $\varepsilon$ respectively.

\subsection{Overview of the Proposed Framework}
Figure~\ref{framework} illustrates the details of our framework. We propose a Spectrum-Enhanced and Environment-Constrainted Graph Fraud Detector (SEC-GFD) to comprehensively integrate rich spectrum and label information into a fraud detector. In particular, SEC-GFD first decomposes the spectrum of the graph signal. Then it performs complicated message passing technique based on these frequency bands respectively to learn comprehensive representations. Then we design a local environmental constraint module to iteratively enhance the high-order associations between nodes' true labels and their surrounding neighbor environments.

\subsection{Hybrid-pass Filter Module}

In the process of designing a hybrid band-pass filter, for purpose of ensuring the division of the spectrum into appropriate frequency bands, we utilize BWGNN~\cite{ref24} as our backbone. According to this, the beta wavelet transform kernel function can be written as:

\begin{equation}
    \mathcal{W}_{p,q}=\boldsymbol{U}\beta_{p,q}^*(\boldsymbol{\Lambda})\boldsymbol{U}^T=\frac{(\frac L2)^p(I-\frac L2)^q}{2B(p+1,q+1)},
\end{equation}

\noindent in which $p+q=C$. Then we decompose the spectrum into $(C+1)$ frequency bands in the frequency domain, which is demonstrated below:

\begin{equation}
    \mathcal{W}=(W_{0,C},W_{1,C-1},\cdotp\cdotp\cdotp,W_{C,0}).
\end{equation}

It is observed that $\mathcal{W}_{p,q}$ is actually a C power polynomial, and $C$ represents how many hops of neighbors' information are considered. Since the Laplacian matrix measures the differences of nodes, it can overcome the over-smoothing problem of vanilla GNN to a certain extent and can be extended to higher order neighbors. When unpacking $\mathcal{W}_{p,q}$, $W_{C,0}$ denotes the C power of the Laplacian matrix $L$. The remained polynomials are the couplings of high-frequency information of neighbors below order C and low-frequency information of low-order neighbors. Furthermore, pure high-frequency information of low-order neighbors is lacked in these polynomials, thus we expand the corresponding part and design a hybrid-pass filter. When adding the pure high-frequency information from order 1 to $(C-1)$, we use $\varepsilon I-D^{-1/2}AD^{-1/2}$ or $\varepsilon I - \tilde{A}$~\cite{ref29} as the basis, where $\varepsilon$ is a scaling hyperparameter bounded in $[0,1]$. The low-order neighbor high-frequency filtering transformation is expressed as follows:

\vspace{-0.2cm}
\begin{equation}
\begin{split}
    R_1: \mathcal{F}_H^1 &=\varepsilon I- \tilde{A}=(\varepsilon-1)I+L, \\
    R_2: \mathcal{F}_H^2 &=(\varepsilon I- \tilde{A})^2=((\varepsilon-1)I+L)^2, \\
    \vdots \\
    R_{C-1}:\mathcal{F}_H^{C-1}&=(\varepsilon I- \tilde{A})^{C-1}=((\varepsilon-1)I+L)^{C-1}. \\
\end{split}
\end{equation}

These $(C-1)$ frequency bands identify the clean high-frequency information of the low-order neighbors $\mathcal{R}=(R_1,R_2,$ $\cdotp\cdotp\cdotp,R_{C-1})$. So far, we obtain 2C frequency bands, consisting of $(C+1)$ band-pass frequency bands and $(C-1)$ high-pass frequency bands for low-order neighbors. Thus we obtain the $Hybrid$ spectral combination:

\begin{equation}
\begin{split}
    &Hybrid = Concat(\mathcal{W},\mathcal{R}) \\
    &= (\overbrace{W_{0,C},W_{1,C-1}, \cdotp\cdotp\cdotp,W_{C,0}}^{\text{(C+1) Band-pass}}, \overbrace{R_1,\cdotp\cdotp\cdotp,R_{C-1}}^{\text{(C-1) High-pass}}).
\end{split}
\end{equation}

So far there are a total of $2C$ frequency bands. Then message transmission function is carried out on 2C frequency bands separately, and finally the learned representations of each frequency band are aggregated:

\begin{equation}
\begin{split}
    H_0 &= MLP(X), \\
    \mathcal{B}_i &= W_{i,C-i}H_0 \quad \mathcal{H}_j = R_{j}H_0, \\
    H &= f_{agg}(\mathcal{B}_0,\cdotp\cdotp\cdotp, \mathcal{B}_C, \mathcal{H}_1,\cdotp\cdotp\cdotp, \mathcal{H}_{C-1}),
\end{split}
\end{equation}

\noindent where $MLP$~\cite{ref30} denotes a simple multi-layer perceptron and $f_{agg}$ can be any aggregation function~\cite{ref8} such as summation or concatenation. $H$ is the aggregated representation of hybrid-pass filter, then weighted cross-entropy loss~\cite{ref31} is used for the training of hybrid-pass filter module:

\begin{equation}
    \mathcal{L}_{hybrid}=\sum_{v\in\mathcal{V}}\left[\delta y_v\log p_v+(1-y_v)\log{(1-p_v)}\right],
\end{equation}

\noindent where $\delta$ is the ratio of anomaly labels $(y_v = 1)$ to normal labels $(y_v = 0)$ in the training set.

\subsection{Local Environmental Constraint Module}

In order to improve the label utilization, and to fully consider the structure and context information of local environments, we add a local environmental constraint module. Specifically, we first mask the characteristics of the target node to prevent the influence of the characteristics of the node itself. Afterwards vanilla GNN is used to jointly learn the feature information of the target node's multi-hop neighbors and then aggregate them together. Obviously, since the features of the target nodes are masked, thus let $h_t^{(0)}=0$. The aggregation function is as follows:

\begin{equation}
    h_t^{(l+1)}=\text{UPDATE}\left(h_t^{(l)},\text{AGG}\left(\{h_v^{(l)}:v\in\mathcal{N}_t\}\right),\right. 
\end{equation}

\begin{equation}
    H_{t}^{neigh} = \text{AGG}(h_{t}^{1}, h_{t}^{2},..., h_{t}^{L}),
\end{equation}

\noindent where $H_{t}^{neigh}$ is the final neighbor aggregation representation. For purpose of exploring the essential association between neighbors with different hop counts, a simple SGC~\cite{ref32} is used without adding any nonlinear functions. Moreover, in order to alleviate the over-smoothing effect~\cite{ref33}, $L$ is generally selected as 2 or 3 in the experiments.

Then we progress graph augmentation and construct KNN view of fraud graph according to the features. Concretely, the k-nearest neighbors~\cite{ref34} of each node are found according to the feature cosine similarity. To ensure the rationality of local environmental features, we do not add any nonlinear functions, but employ the average pooling operation.

\begin{equation}
    H_{t}^{knn}=f_{mean}(\{\mathbf{x}_{u}|\forall u\in K_{t}\})=\frac{1}{|K_{t}|}\sum_{u\in K_{t}}\mathbf{x}_{u}.
\end{equation}

So far, we have obtained the multi-hop neighbor information of the target nodes and their k-nearest neighbor environmental feature information. In order to realize the local environmental constraints, this paper judges based on common senses that, the multi-hop neighbor features learned by normal nodes and the local environmental features should be as similar as possible, while the two features learned by anomaly nodes should have a large deviation. Following the formula of InfoNCE loss~\cite{ref35} in Contrastive Learning, we design a constraint loss function $\mathcal{L}_{env}$ specially adapted to the above judgments:

\begin{equation}
    \mathcal{L}_{env} = -log\frac{\frac{1}{\left | \mathcal{V}_n \right | } \sum_{v_i\in \mathcal{V}_n}^{} e_{v_i}^{sim(H_{v_i}^{neigh}, H_{v_i}^{knn})} }{\frac{1}{\left | \mathcal{V}_a \right | } \sum_{v_j \in \mathcal{V}_a}^{} e_{v_j}^{sim(H_{v_j}^{neigh}, H_{v_j}^{knn})} } ,
\end{equation}

\noindent where $\mathcal{V}_n$ and $\mathcal{V}_a$ represent the set of normal nodes and anomaly nodes in the training set respectively, and $sim(\cdot ,\cdot )$ is the simple cosine similarity operator.

\subsection{Fraud Detection}
For the sake of jointly learning hybrid frequency-band representations and applying local environmental constraints, we carefully design the objective function of SEC-GFD. Specifically, we construct a weighted binary cross-entropy loss function and an InfoNCE-like contrastive loss to realize the constraints of the surrounding environments on the target nodes. The final loss function is as follows:

\begin{equation}
    \mathcal{L} = \alpha \mathcal{L}_{hybrid} + (1-\alpha) \mathcal{L}_{env}.
\end{equation}

\noindent In order to better combine the two loss functions, a hyperparameter $\alpha \in [0,1]$ is added to balance their influences.

\section{Experiments}

In this section, we evaluate the performance of our proposed approach on real-world datasets through a series of experiments, and compare the results with those of state-of-the-art baselines to demonstrate the effectiveness of our approach. In addition, we conduct experiments on the response of heterophily edges to distinct filters, as well as experiments on ablation and hyperparameter analysis.

\vspace{-0.2cm}
\subsection{Experimental Settings}
\textbf{Dataset.} We conduct experiments on four real-world datasets targeted at fraud detection scenarios. Overall, they are Amazon~\cite{ref36}, YelpChi~\cite{ref37}, and two recently published transaction datasets, i.e., T-Finance and T-Social~\cite{ref24}. The YelpChi dataset consists of filtered and recommended hotel and restaurant reviews from Yelp. The Amazon dataset includes product reviews in the music instrument category. The T-Finance dataset and T-Social dataset detect anomalous accounts in transaction networks and social networks respectively, and they share the same 10-dimensional features. The T-Social dataset is a large-scale dataset, several tens to hundreds of times larger in scale compared to the other datasets. The statistics of these datasets are summarized in Table~\ref{data}.

\begin{table}[!t]
  \centering
  \footnotesize
  \begin{tabular}{c|cc p{1.3cm}<{\centering} p{0.8cm}<{\centering}}
    \toprule
    Dataset & Nodes & Edges & Anomaly(\%) & Features \\
    \midrule
    Amazon & 11,944 & 4,398,392 & 6.87 & 25 \\
    YelpChi & 45,954 & 3,846,979 & 14.53 & 32 \\
    T-Finance & 39,357 & 21,222,543 & 4.58 & 10 \\
    T-Social & 5,781,065 & 73,105,508 & 3.01 & 10 \\
  \bottomrule
\end{tabular}
\caption{Statistics of four datasets.}
\label{data}
\end{table}

\begin{table*}[h]
  \begin{tabular}{c|c|p{1.4cm}<{\centering} p{0.8cm}<{\centering}|p{1.4cm}<{\centering} p{0.8cm}<{\centering}|p{1.4cm}<{\centering} p{0.8cm}<{\centering}|p{1.4cm}<{\centering} p{0.8cm}<{\centering}}
    \toprule
    \multicolumn{2}{c|}{\multirow{3}{*}{Methods}} & \multicolumn{8}{c}{Datasets} \\
    \cline{3-10}
    \multicolumn{2}{c|}{} & \multicolumn{2}{c|}{YelpChi}  & \multicolumn{2}{c|}{Amazon} 
    & \multicolumn{2}{c|}{T-Finance} & \multicolumn{2}{c}{T-Social}\\
    \multicolumn{2}{c|}{} & F1-macro & AUC & F1-macro & AUC & F1-macro & AUC & F1-macro & AUC \\
    \hline
    \multirow{4}{*}{Homophily GNN} & GCN & 53.21 & 56.80 & 63.52 & 80.14 & 71.47 & 66.31 & 61.73 & 87.84  \\
    & GraphSAGE & 64.34 & 73.61 & 75.25 & 86.73 & 55.27 & 69.52 & 60.49 & 72.65 \\
    & GAT & 55.86 & 58.92 & 82.41 & 89.73 & 54.27 & 75.29 & 73.72 & 88.61  \\
    & GIN & 63.42 & 75.28 & 71.14 & 81.38 & 65.72 & 81.63 & 63.38 & 81.33  \\
    \hline
    \multirow{4}{*}{GFD Algorithm} & GraphConsis & 57.91 & 69.55 & 78.46 & 87.27 & 73.58 & 91.42 & 58.32 & 73.89  \\
    & CARE-GNN & 63.58 & 79.42 & 83.31 & 91.57 & 65.37 & 91.93 & 55.92 & 68.78 \\
    & PC-GNN & 64.75 & 77.14 & 91.73 & 95.63 & 77.25 & 92.83 & 57.21 & 71.27 \\
    & GAGA & \underline{76.71} & 89.77 & 90.31 & 95.61 & 85.13 & 94.72 & 76.64 & 87.93  \\
    \hline   
    \multirow{4}{*}{Heterophily GNN} & ACM & 69.72 & 88.28 & 81.83 & 93.69 & 85.48 & \underline{96.02} & 81.27 & 90.79  \\
    & H2-FDetector & 69.38 & 88.63 & 83.84 & 96.41 & \underline{87.39} & 95.41 & 78.89 & 88.56 \\
    & BWGNN & 76.66 & 90.56 & \underline{91.68} & \underline{97.52} & 84.59 & 95.29 & \underline{84.58} & \underline{95.27}  \\
    & GDN & 76.05 & \underline{90.79} & 90.68 & 97.09 & 86.12 & 94.28 & 80.63 & 89.35 \\
    \hline
    ours & SEC-GFD & \textbf{77.73} & \textbf{91.39} & \textbf{92.35} & \textbf{98.23} & \textbf{89.86} & \textbf{96.32} & \textbf{87.74} & \textbf{96.11} \\
  \bottomrule
\end{tabular}
\caption{Fraud detection performance on four real-world datasets in terms of F1-macro and AUC value. (The bold and italicized values in the table represent the best and second-best performance, respectively.)}
\label{results}
\end{table*}


\noindent \textbf{BaseLines for Comparison.} We conduct comparative experiments between SEC-GFD and advanced baselines that belong to three groups. The first group consists of methods based on homophily GNN models, including GCN~\cite{ref10}, GraphSAGE~\cite{ref11}, GAT~\cite{ref12} and GIN~\cite{ref13}. The second group comprises state-of-the-art methods for graph-based fraud detection, namely GraphConsis~\cite{ref16}, CARE-GNN~\cite{ref17}, PC-GNN~\cite{ref18} and GAGA~\cite{ref19}. The third group contains GNN-based fraud detection models specifically designed for heterophily, including ACM~\cite{ref22}, H2-FDetector~\cite{ref23}, BWGNN~\cite{ref24} and GDN~\cite{ref40}. More details on description and implementation of the benchmarks are provided in Appendix B.

\noindent \textbf{Metrics.} As graph anomaly detection poses a class-imbalanced classification problem, this paper utilizes two widely adopted metrics: F1-macro and AUC. F1-macro considers the weighted average of F1 scores across multiple classes, and AUC is the area under the ROC Curve.

\noindent \textbf{Settings.} In our proposed method SEC-GFD, for the Amazon, YelpChi, and T-Finance datasets, the hidden layer dimension is set to 64, and the high-frequency signal neighbor order C is set to 2. For the T-Social dataset, the hidden layer dimension is set to 16, the order C of high-frequency signal neighbors is set to 5. The experimental results show that, for such a large-scale graph, fusing high-frequency information of higher-order neighbors can learn better features. In the experiments, to ensure fairness, the size of training/validation/testing set of the datasets is set to 0.4/0.2/0.4 for all the compared methods. Our method is implemented with the DGL library in Pytorch, and for other methods, public source codes are utilized for implementation. Furthermore, we run for 100 epochs on the four datasets for all methods.

\vspace{-0.2cm}
\subsection{Experimental Results}

We conduct a comprehensive comparison of our approach with vanilla homophily GNN models, state-of-the-art graph-based fraud detection models and novel GNNs targeted at heterophily. The corresponding results are shown in Table~\ref{results}.

First of all, It is observed that SEC-GNN outperforms all vanilla homophily GNN models, \ie, GCN, GraphSAGE, GAT and GIN. For instance, when compared with GIN, SEC-GFD achieves improvements of 13.37\% and 26.41\% in terms of F1-macro and AUC on Amazon. To delve reasons, these models can be regarded as low-pass filters in essence and only focus on low frequency information. While our SEC-GNN integrates rich information of distinct frequency bands adapting to heterophily.

When compared with state-of-the-art graph-based fraud detection methods, these methods still have a large gap with our approach. Among them, GraphConsis demonstrates the worst comprehensive performance. It combines the characteristics of nodes with the surrounding environments to learn relationship attention weights, and it is essentially a low-pass filter. CARE-GNN and PC-GNN achieve some improvements in the effectiveness. This is because they are aware of heterophily edges and camouflage problems, and design corresponding neighbor selection module. GAGA introduces a group aggregation module to generate distinguishable multi-hop neighborhood information. Although it demonstrates substantial overall improvements, it still exhibits a significant disparity when compared to our proposed method. Specically, when compared with GAGA, SEC-GFD achieves gains of 2.04\% and 2.62\% in terms of F1-macro and AUC on YelpChi. The reason behind this can be attributed to that our method additionally learns complicated high-frequency information in addition to learning multi-hop neighbor information.

Finally, when compared with novel heterophily GNNs, experimental results in Table~\ref{results} show that our proposed method consistently outperforms them on all metrics across the four datasets. To be specific, SEC-GFD attains improvements of 2.47\% and 1.71\% in terms of F1-macro and AUC on T-Finance. As to T-Social, SEC-GFD achieves gains of 3.78\% and 0.84\% in terms of F1-macro and AUC. Among these models, ACM studies heterophily in terms of feature similarity, H2-FDetector identifies homophily and heterophily connections and applies different aggregation strategies, BWGNN designs a band-pass filter to transmit information in multiple frequency bands, and GDN divides the node features into anomaly pattern and neighborhood pattern. The superior performance of SEC-GFD indicates the effectiveness of the proposed hybrid filtering module and local environmental constraint module.

\vspace{-0.2cm}
\subsection{Heterophily Edges Clipping Analysis}

\begin{figure}[htbp]
	\centering
	\includegraphics[width=0.47\textwidth]{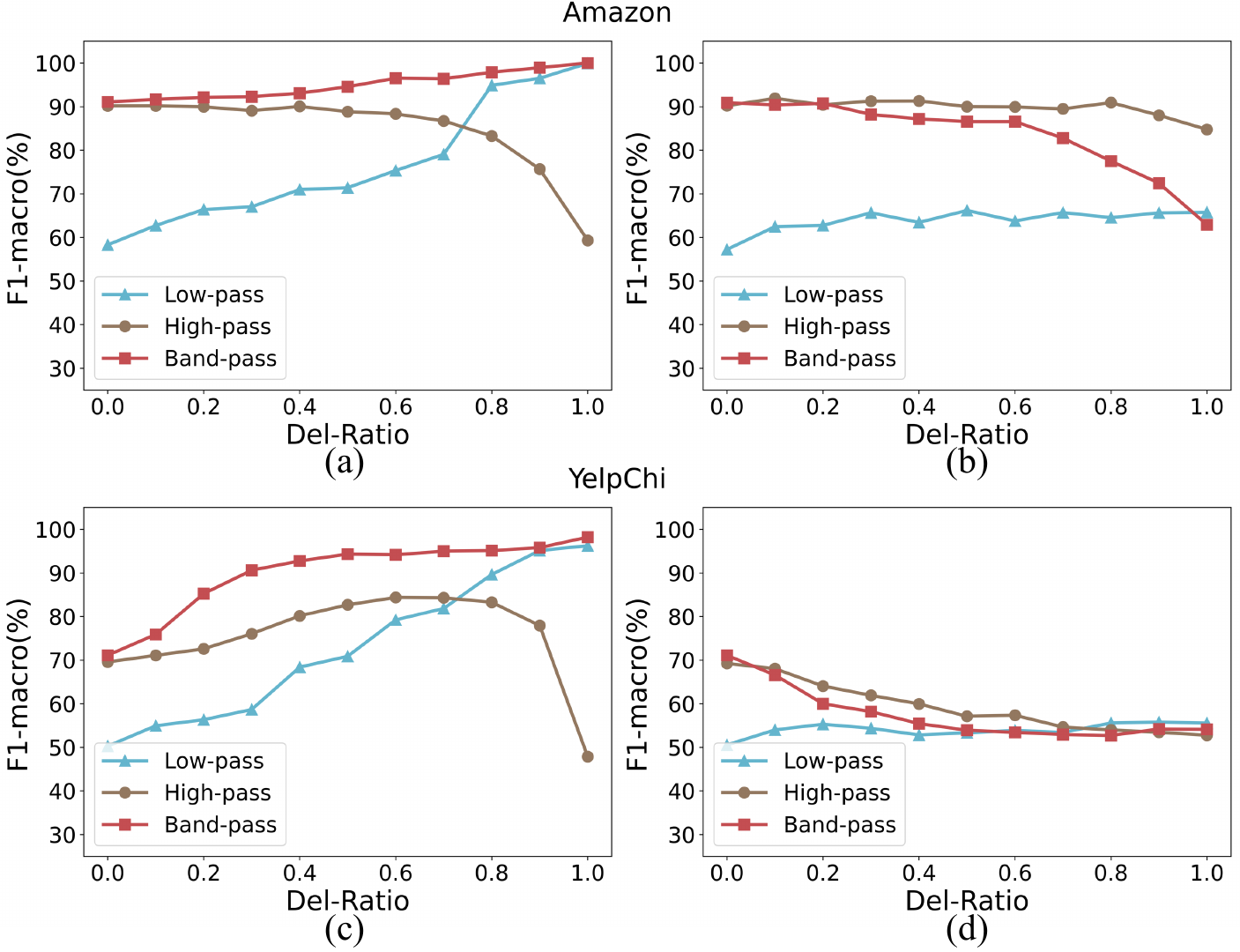}
	\caption{Filters' performance with different heterophily edges' deletion ratio on Amazon and YelpChi. Fig (a) and Fig (c) denote deletion of heterophily edges on the entire graph, while Fig (b) and Fig (d) denote deletion of heterophily edges appearing in training graph.}
	\label{fig:del}
\end{figure}

To explore the effect of heterophily edge pruning on GNNs with different filter properties, we conduct extensive experiments. In this paper, three message passing mechanisms with low-pass, high-pass, and band-pass filtering characteristics are designed for experiments. The low-pass and high-pass filters use the adjacency matrix and normalized Laplacian matrix respectively for message passing, and the band-pass filter uses the beta wavelet filter function from BWGNN, with two layers of neighbor information are considered. The experimental results on Amazon and Yelpchi are illustrated in Figure~\ref{fig:del}.


Considering the difficulty in discriminating heterophily edges, in many cases only some heterophily edges connected with labeled nodes on the training set can be identified. Thus two sets of experiments are conducted in this paper. As observed in Figure~\ref{fig:del}, when clipping the heterophily edges of the whole graph, the performance of low-pass and band-pass filtering improves with the ratio of heterophily edge clipping. While the performance of high-pass filtering shows a monotonous decrease. And the performance of the band-pass filter keeps the best. Moreover, when clipping heterophily edges appearing in the training set, results are completely different. Specifically, low-pass filter maintains a poor performance, while band-pass and high-pass filter exhibit different degrees of declining trends. This suggests that indiscriminate removal of heterophily edges is not necessarily beneficial.

\vspace{-0.2cm}
\subsection{Effectiveness of Each Component of SEC-GFD}
To explore how each component contributes to the performance of SEC-GFD, we systematically remove certain parts of the framework for ablation study. The results are illustrated in Table~\ref{ablation}. 

Firstly, in order to explore the performance of the two parts of the hybrid filter, we omit band-pass bands and low-order pure high-pass bands respectively. When band-pass bands are excluded, the performance drops by 2.06\% and 2.67\% in terms of AUC and F1-macro. And the performance decreases 1.13\% and 1.27\% in terms of AUC and F1-macro when low-order high-pass bands are removed. It is obviously that band-pass bands are more important, this is because band-pass bands consider high frequency and lower frequency information together. Afterwards, we delete the local environmental constraint, then performance drops by 0.86\% and 1.31\% in terms of AUC and F1-macro. However, it still outperforms other state-of-the-art methods, this also reflects the effectiveness of our hybrid filter. Thus the hybrid filter can be equipped as a general backbone to various other GNN-based fraud detectors to improve performance.


\begin{table}[t]
\centering
\small
  \begin{tabular}{p{2.1cm}<{\centering}|p{1.3cm}<{\centering}|p{0.7cm}<{\centering}|p{1.3cm}<{\centering}|p{0.7cm}<{\centering}}
    \toprule
    & \multicolumn{2}{c}{Amazon} & \multicolumn{2}{c}{T-Finance} \\
    &  F1-macro & AUC & F1-macro & AUC \\
    \midrule
    SEC-GFD & \textbf{92.27} & \textbf{98.19} & \textbf{89.74} & \textbf{96.34} \\
    w/o high-pass & 91.48 & \underline{97.28} & \underline{88.65} & 95.05 \\
    w/o band-pass & 90.29 & 96.31 & 86.79 & 93.77 \\
    w/o env       & \underline{91.88} & 97.14 & 88.14 & \underline{95.21} \\
    \bottomrule
  \end{tabular}
  \caption{Ablation Performance.(The bold and italicized values denote the best and runner-up performance.)}
  \label{ablation}
\end{table}

\vspace{-0.2cm}

\subsection{Selection of C for high-order neighbors}

\begin{figure}[!t]
\centering
\includegraphics[width=0.47\textwidth]{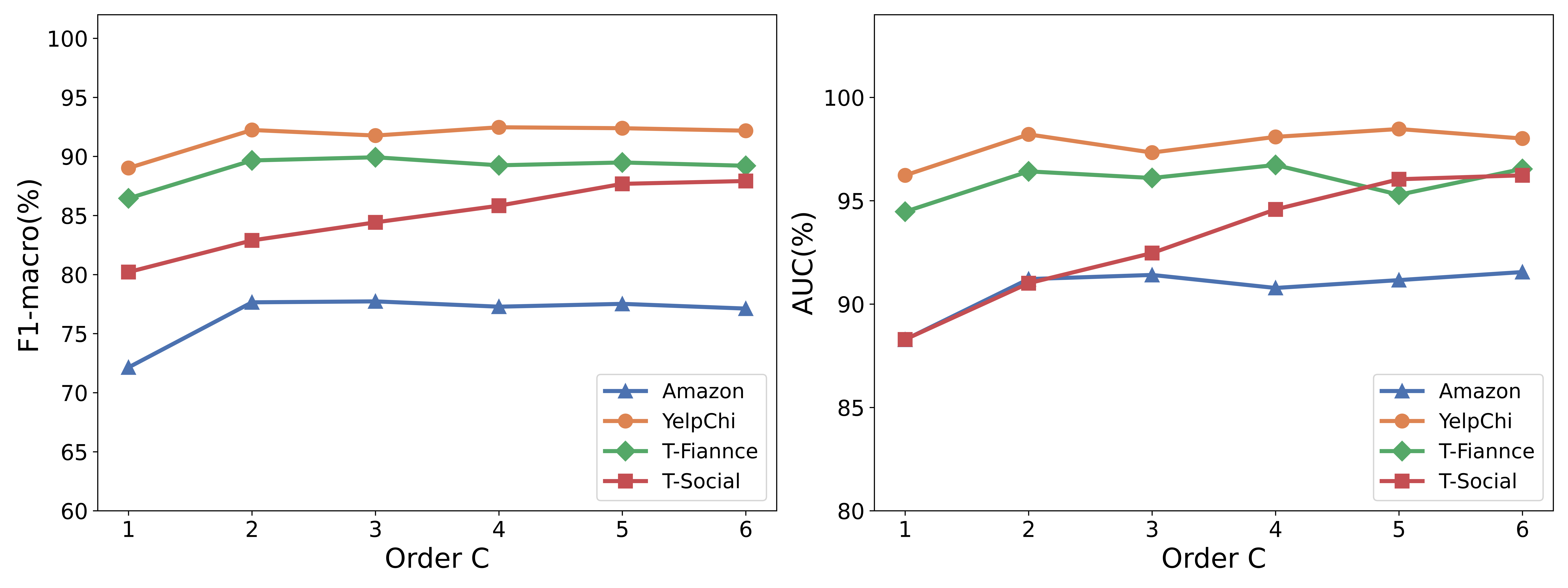}
\caption{Performance with different order C.}
\label{order}
\end{figure}

In the previous discussion, based on the formula of beta wavelet, we judge that C represents the order of neighbor information considered in the network training process. Therefore, the order C is an extremely important parameter. Figure~\ref{order} shows the changes of F1 and AUC values with the changes of C from 1 to 6 on four real-world datasets.

It is observed that on the three smaller datasets, the improvement of the effect is no longer obvious when C exceeds two. While on large-scale dataset T-Social, which is nearly a hundred times larger than other datasets, there is still a significant performance improvement until C reaches five. A possible reason is that, larger fraud graph may contain more complicated inter-node interactions and more common spread of fraudulent behavior across multi-hop neighbors.

\section{Conclusion}

This paper reviews the research on highly heterophily graphs, especially in fraud detection scenarios, and further explores the association between heterophily edges and spectral domain energy distribution. The experimental results prove that it is not necessary to discriminate and cut heterophily edges, but to make full use of each frequency band for information transmission to achieve excellent results. To this end, we have designed a spectrum-enhanced and environment-constrainted graph-based fraud detector SEC-GFD. The main body is a hybrid filtering module, which contains multiple frequency bands with multi-order neighbors. Then it is supplemented with a local environmental constraint module to improve the utilization of node labels and the adaptive relationships between nodes and their surrounding environments.

\section{Acknowledgments}

This work is supported in part by the Fundamental Research Funds for the Central Universities under Grant 2022RC026; in part by the CCF-NSFOCUS Open Fund; in part by the NSFC Program under Grant 62202042, Grant U20A6003, Grant 62076146, Grant 62021002, Grant U19A2062, Grant 62127803, Grant U1911401 and Grant 6212780016; in part by Industrial Technology Infrastructure Public Service Platform Project `Public Service Platform for Urban Rail Transit Equipment Signal System Testing and Safety Evaluation' (No. 2022-233-225), Ministry of Industry and Information Technology of China.

\bibliography{aaai24}

\clearpage

\appendix

\section{A. Bond between Rayleigh Quotient and Spectral Energy}

\vspace{0.2cm}

For the Rayleigh quotient, we initially examine its numerator, which corresponds to the dot product of the signal with itself following the Laplacian matrix transformation. Subsequently, we proceed to expand the formula and obtain:

\begin{equation}
\begin{split}
       X^{T}LX &= X^{T}(D - A)X = X^{T}DX - X^{T}AX \\
       &= \sum_{i=1}^{N} x_{i}^2 d_i - \sum_{i=1}^{N} \sum_{j=1}^{N} A_{ij} x_{i}x_{j} \\
       &= \frac{1}{2} (\sum_{i=1}^{N} x_{i}^2 d_i - 2\sum_{i=1}^{N} \sum_{j=1}^{N} A_{ij} x_{i}x_{j} + \sum_{j=1}^{N} x_{j}^2 d_j) \\
       &= \frac{1}{2} \sum_{i=1}^{N}\sum_{j=1}^{N}A_{ij}(x_{i}-x_{j})^2
\end{split}
\nonumber  
\end{equation}

\vspace{-0.2cm}

\noindent Upon observation, we note that the numerator is similar to a weighted sum of Euclidean distances, which can also serve as a metric for signal dissimilarity. Subsequently, we substitute this into the formula of the Rayleigh quotient:

\begin{equation}
\begin{split}
\noindent
    R(L,X) = \frac{X^{T}LX}{X^{T}X} 
    &= \frac{\sum_{i=1}^{N}\sum_{j=1}^{N}A_{ij}(x_{i}-x_{j})^2}{2 \sum_{i=1}^{N} x_{i}^2} \\
    &= \frac{\sum_{(i,j) \in \mathcal{E}}^{} (x_{i}-x_{j})^2 }{2\sum_{i \in \nu }^{}x_{i}^{2}}  
\end{split}
\nonumber
\end{equation}

\vspace{0.2cm}

\noindent Thus Rayleigh quotient is actually the standardized variance fraction of signal X. To establish a correlation with spectral energy, the analysis is conducted in the spectral domain. According to spectral graph theory, $X^{T}LX=\sum_{i=1}^{N} \lambda_{i} \tilde{x}_{i}^2$ and $X^{T}X=\sum_{i=1}^{N} x_{i}^2$ have been empirically validated, thus the Rayleigh quotient can also be expanded as follows:

\begin{equation}
\begin{split}
    R(L,X) &= \frac{X^{T}LX}{X^{T}X} 
    = \frac{\sum_{i=1}^{N} \lambda_{i} \tilde{x}_{i}^2}{\sum_{i=1}^{N} x_{i}^2} \\
    &= \sum_{i=1}^{N} \lambda_{i}(\frac{\tilde{x}_{i}^2}{\sum_{k=1}^{N}\tilde{x}_{k}^2}) \\
    &= \sum_{i=1}^{N} \lambda_{i}(\eta_{i} - \eta_{i-1}) \\
    &= \lambda_{N} - \left [(\lambda_1-\lambda_0)\eta_0+\cdots + (\lambda_N-\lambda_{N-1})\eta_{N-1} \right ] \\
    &= \lambda_{N} - \left [ \int_{0}^{\lambda_1}f(t)d_t + \cdots + \int_{\lambda_{N-1}}^{\lambda_{N}}f(t)d_t \right ] \\
    &= \lambda_{N} * f(t)_{max} - S_{spec}
\end{split}
\nonumber
\end{equation}

\vspace{-0.2cm}

\noindent where $S_{spec}$ is the area under spectrum energy accumulation curve and $\lambda_{N}$ represents the total length of the continuous spectrum. Consequently, the Rayleigh quotient exhibits a higher value when the proportion of high-frequency components is greater.

\section{B. Baselines and Implementation Details}

Baselines can be classified into three distinct categories. The first category contains GNNs based on homophily assumption:
\begin{itemize}
    \item \textbf{GCN} (Kipf and Welling 2016): GCN is a traditional graph neural network model that aggregates neighbor information.
    \item \textbf{GraphSAGE} (Hamilton, Ying, and Leskovec 2017): GraphSAGE is a graph neural network model that samples and aggregates information from node neighborhoods to learn node representations.
    \item \textbf{GAT} (Velickovic et al. 2017): GAT is a graph neural network model that applies attention mechanisms to process graph data.
    \item \textbf{GIN} (Xu et al. 2018): GIN is a graph neural network model that aggregates node information using permutation-invariant operations.
\end{itemize}

The second category are state-of-the-art graph-based fraud detection methods:
\begin{itemize}
    \item \textbf{GraphConsis} (Liu et al. 2020): GraphConsis combines the characteristics of nodes with the surrounding environment to learn a relationship attention weight.
    \item \textbf{CARE-GNN} (Dou et al. 2020): CARE-GNN achieves the effect of modifying the adjacency matrix and pruning the edges by designing the neighbor selection module.
    \item \textbf{PC-GNN} (Liu et al. 2021): PC-GNN introduces a node sampler and a label-aware neighbor selector to further reweight imbalanced classes.
    \item \textbf{GAGA} (Wang et al. 2023): GAGA introduces a group aggregation module to generate distinguishable multi-hop neighborhood information.
\end{itemize}

The third category consists of GNN methods specifically designed for heterophily:
\begin{itemize}
    \item \textbf{ACM} (Luan et al. 2022): ACM uses aggregated features to study heterophily in terms of feature similarity.
    \item \textbf{H2-FDetector} (Shi et al. 2022): H2-FDetector identifies homophily and heterophily connections and applies different aggregation strategies.
    \item \textbf{BWGNN} (Tang et al. 2022): BWGNN designs a band-pass filter to transmit information in multiple frequency bands.
    \item \textbf{GDN} (Gao et al. 2023): GDN divides the node features into two parts that describe the abnormal pattern and adapt to the neighborhood information.
\end{itemize}

\end{document}